\documentclass[10pt,twocolumn,letterpaper]{article}
\usepackage{iccv}
\usepackage{times}
\usepackage{epsfig}
\usepackage{graphicx}
\usepackage{amsmath}
\usepackage{amssymb}
 \usepackage{array,multirow,graphicx}
\usepackage{arydshln}
\usepackage{siunitx}
\usepackage{pifont}
\usepackage{xcolor}

\DeclareMathOperator*{\argminA}{min\,} 
\DeclareMathOperator*{\argmaxA}{max\,} 

\makeatletter
\@namedef{ver@everyshi.sty}{}
\makeatother

\usepackage[utf8x]{inputenc}
\usepackage{pgfplots}
\usepackage{pgfplotstable}
\pgfplotsset{compat=newest}
\usepackage{tikz}
\usetikzlibrary{backgrounds}
\usepackage{color}
\usepackage{booktabs,siunitx} 
\usepackage{makecell}
\usepackage{enumitem}
\usepackage[caption=false]{subfig}
\usepackage[noadjust]{cite}
\usepackage{rotating}

\newcommand{\nangthirty}{40.398128707029535}
\newcommand{\nangfifty}{31.138505245410684}
\newcommand{\nangeighty}{15.71422196306487}
\newcommand{\nangonetwenty}{12.749144084494912}
\newcommand{\nangtlnear}{75.5175940282943}
\newcommand{\nangtlfar}{24.4824059717057}
\newcommand{\nangcwnear}{61.57933531226115}
\newcommand{\nangcwfar}{38.42066468773884}
\newcommand{\nangleft}{49.58975390953642}
\newcommand{\nangstraight}{12.49768050744374}
\newcommand{\nangright}{37.912565583019834}
\newcommand{\nangenter}{26.308889277562958}
\newcommand{\nanginside}{51.10564265392751}
\newcommand{\nangexit}{12.638439277596596}
\newcommand{\nangnone}{9.947028790912938}
\newcommand{\nvelthirty}{35.53559427118589}
\newcommand{\nvelfifty}{24.17403056073403}
\newcommand{\nveleighty}{20.91181241610571}
\newcommand{\nvelonetwenty}{19.37856275197436}
\newcommand{\nveltlnear}{53.88085716440062}
\newcommand{\nveltlfar}{46.11914283559937}
\newcommand{\nvelcwnear}{53.1749479368665}
\newcommand{\nvelcwfar}{46.82505206313349}
\newcommand{\nvelleft}{34.28794137969079}
\newcommand{\nvelstraight}{28.67206826100369}
\newcommand{\nvelright}{37.03999035930552}
\newcommand{\nvelenter}{27.152222909359946}
\newcommand{\nvelinside}{28.1142809614425}
\newcommand{\nvelexit}{24.803130463996744}
\newcommand{\nvelnone}{19.930365665200817}

\definecolor{30}                    {RGB}{188, 193, 245}
\definecolor{50}                    {RGB}{144, 152, 238}
\definecolor{80}                    {RGB}{100,111,232}
\definecolor{120}                   {RGB}{55,70,225}

\definecolor{Light Near}            {RGB}{255,198,128}
\definecolor{Light Far}             {RGB}{255,152,26}

\definecolor{Crossing Near}         {RGB}{153,230,153}
\definecolor{Crossing Far}          {RGB}{71,210,71}

\definecolor{Left}                  {RGB}{255,128,128}
\definecolor{Straight}              {RGB}{255,77,77}
\definecolor{Right}                 {RGB}{255,26,26}

\definecolor{Enter}                 {RGB}{219, 190, 244}
\definecolor{Inside}                {RGB}{195,146, 236}
\definecolor{Exit}                  {RGB}{170,103, 229}
\definecolor{None}                  {RGB}{146,60, 221}
\iccvfinalcopy 

\def\iccvPaperID{1726} 

\newcommand{\YES}{\ding{51}}

\newcommand*\rot[1]{\rotatebox{90}{#1}}

\ificcvfinal\fi
\begin{document}

\title{Learning Accurate, Comfortable and Human-like Driving}

\author{Simon Hecker\\
Computer Vision Lab\\
ETH Z\"urich\\
{\tt\small heckers@vision.ee.ethz.ch}
\and
Dengxin Dai\\
Computer Vision Lab\\
ETH Z\"urich\\
{\tt\small dai@vision.ee.ethz.ch}
\and
Luc Van Gool\\
Computer Vision Lab\\
ETH Z\"urich\\
{\tt\small vangool@vision.ee.ethz.ch}
}

\maketitle

\begin{abstract}

Autonomous vehicles are more likely to be accepted if they drive accurately, comfortably, 
but also similar to how human drivers would. This is especially true when autonomous and 
human-driven vehicles need to share the same road. The main research focus thus far, however, 
is still on improving driving accuracy only. This paper formalizes the three concerns with the aim of accurate, 
comfortable and human-like driving. Three contributions are made in this paper. 
First, numerical map data from HERE Technologies are employed for more accurate driving; 
a set of map features -- which are believed to be relevant to driving -- are engineered to navigate better. 
Second, the learning procedure is improved from a pointwise prediction to a sequence-based prediction and passengers' 
comfort measures are embedded into the learning algorithm. Finally, we take advantage of the advances in 
adversary learning to learn human-like driving; specifically, the standard L1 or L2 loss is augmented by 
an adversary loss which is based on a discriminator trained to distinguish between human driving and machine driving. 
Our model is trained and evaluated on the Drive360 dataset, which features $60$ hours and $3000$ km of 
real-world driving data. Extensive experiments show that our driving model is more accurate, more 
comfortable and behaves more like a human driver than previous methods. The resources of this work will be 
released on the project page. 

\end{abstract}

\section{Introduction}
\label{sec:intro}

The deployment of autonomously driven cars is imminent, given the advances in perception, robotics and sensor technologies. It has been hyped that autonomous vehicles of multiple companies have driven millions of miles. Yet, their platforms, algorithms and assessment results are not shared with the whole community. It is believed that autonomous vehicles are more likely to be accepted if they drive accurately, comfortably and drive the same way as human drivers would. The rationale is not that human-style driver would somehow be superior, but rather that humans will find it easier to interact and feel at ease with autonomous cars in such case. This concern is especially important for the near future, when autonomous vehicles are new and share the road with human-driven vehicles. Despite its importance, this topic has not received much attention. The current research focus is only on developing more accurate driving methods. This paper formalizes the three concerns into a single learning framework aiming at accurate, comfortable and human-like driving. 

\textbf{Driving Accuracy}. The last years have seen tremendous progress in academia on learning driving models~\cite{end:driving:imitation:18,drive:surroundview:route:planner}. However, many of these systems are deficient in terms of the sensors used, when compared to the driving systems developed by large companies. For instance, many algorithms only use a front-facing camera~\cite{end:driving:16,end:driving:eventcamera:18} (with few exceptions~\cite{drive:surroundview:route:planner}); Maps are exploited only for simple directional commands~\cite{end:driving:imitation:18} or video rendering~\cite{drive:surroundview:route:planner}. While these setups are sufficient to allow the community to study many challenges, developing algorithms for fully autonomous cars requires the use of numerical maps of high fidelity. 

\textbf{Ride Comfort}. Current driving algorithms~\cite{end:driving:16,end:driving:eventcamera:18,end:driving:imitation:18,drive:surroundview:route:planner} mostly treat driving as a regression problem with i.i.d individual training samples, e.g. regressing the low-level steering angle and speed for a given data sample. Yet, driving is a continuous sequence of events over time. Longitudinal and lateral control need to be coupled and these coupled operations need to be combined over time for a comfortable ride~\cite{bezier:curve:path:planning:iv10,combined:longitudinal:lateral:14}. Thus, driving models need to be learned with continuous data sequences and proper passenger comfort measures need to be embedded into the learning system.   
While research on passenger comfort started to receive some attention~\cite{passenger:seat:ride:comfort:15}, it hardly did so in learning driving models. To the best of our knowledge, no published work on learning driving models has incorporated passenger comfort measures. This work minimizes both longitudinal and lateral oscillations, as they contribute significantly to passengers' motion sickness~\cite{motion:sickness:99,passenger:seat:ride:comfort:15}.   

\textbf{Human-like Driving}. It is believed that autonomous vehicles will be better received if they behave like human drivers~\cite{human:like:driving:simulator:01,realistic:driving:03,human:like:motion:planning:17}. During the last years, several human-driven cars crashed into autonomous cars. Thankfully, damages were limited. Although there were too few of these cases yet to clearly reveal the underlying causes, experts believe that part of the problem lies with non-human-like driving behaviours of the autonomous cars. In this work, we take advantage of adversary learning to teach the car about human-like driving. Specifically, a discriminator is trained, together with our driving model, to distinguish between human driving and our machine driving. The driving model is trained to be accurate, comfortable, and at the same time to fool the discriminator so that it believes that the driving performed by our method was by a human driver. A new evaluation criterion is proposed to score the \emph{human-likeness} of a driving model. 


\noindent



This work makes four major contributions: 1) obtaining numerical map data for real-world driving data, engineering map features and showing their effectiveness in a \emph{learning to drive} task; 2) incorporating ride comfort measures into an end-to-end driving framework; 3) incorporating human-like driving style into an end-to-end driving framework; and 4) improving the learning procedure of end-to-end driving from pointwise predictions to sequence-based predictions. As a result, we formalize the three major concerns about autonomous driving into one learning framework. The result is experimentally demonstrated be a more accurate, comfortable and human-like driving model. 



\section{Related Work}


\noindent
\textbf{Learning Driving Models}.
Significant progress has been made in autonomous driving in the last few years. Classical approaches require the recognition of all driving-relevant objects, such as lanes, traffic signs, traffic lights, cars and pedestrians, and then perform motion planning, which is further used for final vehicle control~\cite{AD:Boss:08}. These type of systems are sophisticated, represent the current state-of-the-art for autonomous driving, but they are hard to maintain and prone to error accumulation over the pipeline. 

End-to-end mapping methods on the other hand construct a direct mapping from the sensory input to the maneuvers. The idea can be traced back to the 1980s~\cite{network:autonomous:1980}. Other more recent end-to-end examples include~\cite{LeCun:driving:05,nvidia:driving:16,end:driving:16,lidar:end:driving:18,end:driving:eventcamera:18,end:driving:imitation:18,drive:surroundview:route:planner,E2E:auxiliary:aaai18}. In \cite{nvidia:driving:16}, the authors trained an end-to-end method with a collection of front-facing videos. The idea was extended later on by using a larger video dataset~\cite{end:driving:16}, by adding side tasks to regularize the training~\cite{end:driving:16,E2E:auxiliary:aaai18}, by introducing directional commands~\cite{end:driving:imitation:18} and route planners~\cite{drive:surroundview:route:planner} to indicate the destination, by using multiple surround-view cameras to extend the visual field~\cite{drive:surroundview:route:planner}, by adding synthesized off-the-road scenarios~\cite{chauffeurnet:18}, and by adding modules to predict when the model fails~\cite{driving:failure:prediction}. The main contributions of this work, namely using numerical map data, incorporating ride comfort measures, and rendering human-like driving in an end-to-end learning framework, are complementary to all methods developed before.   

There are also methods dedicated to robust transfer of driving policies from a synthetic domain to the real world domain~\cite{driving:policy:transfer:18,drive:sim2real:19}. Some other works study how to better evaluate the learned driving models~\cite{challenges:AV:testing:16,offline:evaluation:driving:18}. Those works are complementary to our work. Other contributions have chosen the middle ground between traditional pipe-lined methods and the monolithic end-to-end approach. They learn driving models from compact intermediate representations called affordance indicators such as \emph{distance to the front car} and \emph{existence of a traffic light}~\cite{deep:driving,conditional:affordance:learning:18}. Our engineered features from numerical maps can be considered as some sort of affordance indicators. Recently, reinforcement learning for driving has received increased attention~\cite{reinforcement:learning:driving,DRL:dirivng:17,drive:in:a:day:19}. The trend is especially fueled by the release of multiple driving simulators~\cite{AirSim:17,CARLA:simulator}.  

\textbf{Navigation Maps}. Increasing the accuracy and robustness of self-localization on a map~\cite{HD:map:10,HD:map:12,traffic:rules:AV:15} and computing the fastest, most fuel-efficient trajectory from one point to another through a road network~\cite{Driving:knowledge:world:11,route:planning,scenic:driving:route:13,personalized:TripPlanner:15,towards:personalized:routing:15} have both been popular research fields for many years. By now, navigation systems are widely used to aid human drivers or pedestrians. Yet, their integration for learning driving models has not received due attention in the academic community, mainly due to limited accessibility~\cite{drive:surroundview:route:planner}.  We integrate industrial standard numerical maps -- from HERE Technologies -- into the learning of our driving models. We show the advantage of using numerical maps and further combine the engineered features of our numerical maps with the visually rendered navigation routes by~\cite{drive:surroundview:route:planner}.

\textbf{Ride Comfort}. Cars transport passengers. This has led to passenger comfort research for human-driven vehicles~\cite{passenger:comfort:public:14}. Driver comfort is also considered when developing the control system of human-driven vehicles~\cite{autonomous:vehicle:PID:12,Amer2017}. Autonomous cars can lead to concerns about how well-controlled such a car is~\cite{ORVideoGaze}, motion sickness~\cite{passenger:seat:ride:comfort:15} and apparent safety~\cite{passenger:seat:ride:comfort:15}. 
While research on passenger comfort started to receive more attention~\cite{passenger:seat:ride:comfort:15}, it is still missing in current driving models. To address this problem, this work incorporates passenger comfort measures into learned autonomous driving models. 


\textbf{Human-like Driving}. A large body of work has studied human driving styles~\cite{vehicle:corpora:driver:behavior:11,driving:style:survey:15}. Statistical approaches were employed to evaluate human drivers and to suggest improvements~\cite{driving:behavior:smartphones:12,driving:behavior:evaluation:13}. This line of research inspired us to ask whether \emph{one can learn and improve machine driving behaviour such that it is very human-like?}. Human-like driving is hard to quantify. Fortunately, recent advances in adversarial learning provide the tools to extract the \emph{gist} of human-like driving, using it to adjust machine driving so that it becomes more human-like. Some work has studied human-like motion planning of autonomous cars, but it was constrained to simulated scenarios~\cite{human:like:driving:simulator:01,realistic:driving:03}. The closest work to ours was done by Kuderer et al.~\cite{driving:styles:AV:15} where a set of manually-crafted features are used to characterize human driving style. Our method learns the features directly from the data using adversarial neural networks.    
  
 
\section{Approach}

 In this section we describe our contributions to improve end-to-end driving models: in terms of driving accuracy, rider comfort, and human-likeness. 
 

\subsection{Accurate Driving}
\label{sec:basic:model}

End-to-end driving has allowed the community to develop promising driving models based on camera data~\cite{end:driving:16,end:driving:imitation:18,drive:surroundview:route:planner}. The focus has mainly been on perception, not so much navigation. Thus far, the representations for navigation are either primitive directional commands in a simulation environment~\cite{end:driving:imitation:18,conditional:affordance:learning:18} or rendered videos of planned routes in real-world environments~\cite{drive:surroundview:route:planner}. 

We contribute 1) augmenting real-world driving data with numerical map data from HERE Technologies; and 2) designing map features believed to be relevant for driving and integrating them into a driving model. To the best of our knowledge, this work is the first to introduce large-scale numerical map data to driving models in real-world scenarios. Data acquisition and feature design are discussed in Sec.~\ref{sec:here:features}, and  the usefulness of the map data is validated in Sec.~\ref{sec:experiment}.


We adopt the driving model developed in~\cite{drive:surroundview:route:planner}. Given the video $\mathrm{I}$ , the map information $\mathrm{M}$, and the vehicle's location $\mathrm{L}$, a deep neural network is trained to predict the steering angle $s$ and speed $v$ for a future time step. 
All data inputs are synchronized and sampled at the same sampling rate $f$, meaning the vehicle makes a driving decision every $1/f$ seconds. The inputs and outputs are represented in this discretized form. 

We use $t$ to indicate the time stamp, such that all data can be indexed over time. For example, $I_t$ indicates the current video frame and $v_t$ the vehicle's current  speed. Similarly,  $I_{t-k}$ is the $k^{th}$ previous video frame and $s_{t-k}$ is the $k^{th}$ previous steering angle. Since predictions need to rely on data of previous time steps, we denote the $k$ recent video frames by $\mathbf{I}_{[t-k+1,t]} \equiv \langle I_{t-k+1}, ..., I_{t}\rangle$, and the $k$ recent map representations by $\mathbf{M}_{[t-k+1,t]} \equiv \langle M_{t-k+1}, ..., M_{t}\rangle$. 
Our goal is to train a deep network that predicts desired driving actions from the visual observations and the planned route. The learning task can be defined as: 
\begin{equation}
F: (\mathcal{I}_{[t-k+1,t]}, \mathcal{M}_{[t-k+1,t]})  \rightarrow \mathcal{S}_{t+1} \times \mathcal{V}_{t+1}
\label{eq:learning:fun1}
\end{equation}
where $\mathcal{S}_{t+1}$ represents the steering angle space and $\mathcal{V}_{t+1}$ the speed space for future time $t+1$.  
$\mathcal{S}$ and $\mathcal{V}$ can be defined at several levels of granularity. We consider the continuous values directly recorded from the car's CAN bus, where $\mathcal{V} = \{V | 0  \leq V \leq 180$ for speed and $\mathcal{S}=\{S | -720  \leq S \leq 720\}$ for steering angle in our case. Here, kilometer per hour (km/h) is the unit of $v$, and degree ($^\circ$) the unit of $s$.  $M_t$ is either a rendered video frame from the TomTom route planner~\cite{drive:surroundview:route:planner}, or the engineered features for the numerical maps from HERE Technologies (defined in Sec.~\ref{sec:here:features}), or the combination of both. 

In order to keep notations concise, we denote the synchronized data $(I,M)$ as $D$. Without loss of generality, we assume our training data to consist of a long sequence of driving data with $T$ frames in total. Then the basic driving model is to learn the prediction function for the steering angle 
\begin{equation}
 \hat{s}_{t+1} \leftarrow f_{\text{st}}( \mathcal{D}_{[t-k+1,t]}),  
\end{equation}
and the velocity
\begin{equation}
 \hat{v}_{t+1} \leftarrow f_{\text{ve}}( \mathcal{D}_{[t-k+1,t]}), 
\end{equation}
with the objective 
\begin{equation}
\label{eq:basic:objective}
    \sum_{t=1}^{T-1} ( | \hat{s}_{t+1} - s_{t+1}| + \lambda |\hat{v}_{t+1} - v_{t+1}| ), 
\end{equation}
where $\hat{s}$ and $\hat{v}$ are predicted values, and $s$ and $v$ are the ground truth values.

The learning under Eq.~\ref{eq:basic:objective} is straightforward and can be implemented with any standard deep network. This objective, however, assumes the driving decisions at each time step are independent from each other. We believe this is an over-simplification because driving decisions indeed exhibit strong temporal dependencies within a relatively short time range.  In the following section, we reformulate the objective by introducing a \emph{ride comfort} and a \emph{human-likeness score} to better model the temporal dependency of driving actions. 

\subsection{Accurate and Comfortable Driving}
\label{sec:comfortable:driving} 

Multiple concepts relating to driving comfort have been proposed and discussed~\cite{passenger:seat:ride:comfort:15,passenger:comfort:public:14}, such as \emph{apparent safety}, \emph{motion comfort (sickness)}, \emph{level of controllability} and \emph{resulting force}. While those are all relevant, some are hard to quantify. We choose to work on \emph{motion comfort}, which is largely influenced by the vehicle's longitudinal and lateral jerk~\cite{human:comfort:in:vehicle:1980,motion:sickness:99,passenger:seat:ride:comfort:15}.   
Due to the short-term predictive nature of most end-to-end driving models, substantial jerking is an inherent problem. Our comfort component aims at reducing jerk by imposing a temporal smoothness constraint on the longitudinal and lateral oscillations, by minimizing the second derivative of consecutive steering angle and speed predictions. 

Before introducing ride comfort and human-like driving, we reformulate Eq.~\ref{eq:basic:objective}. If the number of consecutive predictions that need to be optimized jointly is denoted by $O$, then minimizing Eq.~\ref{eq:basic:objective} is equivalent to minimizing 
\begin{equation}
     \sum_{t=1}^{T-O} \sum_{o=1}^O ( | \hat{s}_{t+o} - s_{t+o}| + \lambda |\hat{v}_{t+o} - v_{t+o}| ).
\end{equation}
Then for every $O$ consecutive frames starting at time $t$, the loss of driving accuracy will be
\begin{equation}
 \mathcal{L}_t^{\text{acc}}  = \sum_{o=1}^O ( | \hat{s}_{t+o} - s_{t+o}| + \lambda |\hat{v}_{t+o} - v_{t+o}|)
\end{equation}. 

We can now present the objective function for accurate and comfortable driving as
\begin{equation} 
\label{eq:confort:driving}
     \sum_{t=1}^{T-O} \sum_{o=1}^O ( \mathcal{L}_t^{\text{acc}} + \zeta_1 \mathcal{L}_t^{\text{com}}),
\end{equation}
where 
\begin{equation}
\begin{split}
    \mathcal{L}_t^{\text{com}} =  \sum_{o=2}^{O-1}  \frac{ |\hat{s}_{t+o-1} -2\hat{s}_{t+o}+\hat{s}_{t+o+1}|}{(1/f)^2} \\ + \lambda \frac{ |\hat{v}_{t+o-1} -2\hat{v}_{t+o}+\hat{v}_{t+o+1}|}{(1/f)^2}
    \end{split}. 
\end{equation}
$\zeta_1$ is a trade-off parameter to balance the two costs. By optimizing under the objective in Eq.~\ref{eq:confort:driving}, consecutive predictions are learned and optimized together for accurate and comfortable driving. 

\subsection{Accurate, Comfortable \& Human-like Driving}
\label{sec:human-like:driving}
If autonomous cars behave differently from human-driven cars, it is hard for humans to predict their future actions. This unpredictability can cause accidents. Also, if a car behaves more the way its passengers expect, the ride will feel more reassuring. Thus, we argue that it is important to design human-like driving algorithms from the very start. Hence, we introduce a \emph{human-likeness} score. The higher the value, the closer to human driving. Since it is hard to manually define what a human driving style is -- as was done for general comfort measures (Sec.~\ref{sec:comfortable:driving}), we adopt adversarial learning to model it. 

An adversarial learning method consists of a generator and discriminator. Our driving model defined in Eq.~\ref{eq:basic:objective} or in Eq.~\ref{eq:confort:driving} is the generator $\mathbf{G}$. We now describe the training objective for the discriminator. For convenience, we name the short trajectories of $O$ frames used in Sec.~\ref{sec:comfortable:driving} as drivelets. Given the outputs of our driving model for a drivelet $\hat{B}_t=(\hat{s}_{t+1},..., \hat{s}_{t+O}, \hat{v}_{t+1},..., \hat{v}_{t+O})$ and its corresponding ground truth from the human driver  $B_t=(s_{t+1},..., s_{t+O}, v_{t+1},..., v_{t+O})$, the goal is to train a fully-connected discriminator $\mathbf{D}$ using the cross-entropy loss to classify the two classes (i.e. machine and human).  
\setlength{\textfloatsep}{10pt}
\begin{figure}[tb]
\centering
\includegraphics[width=0.9\linewidth]{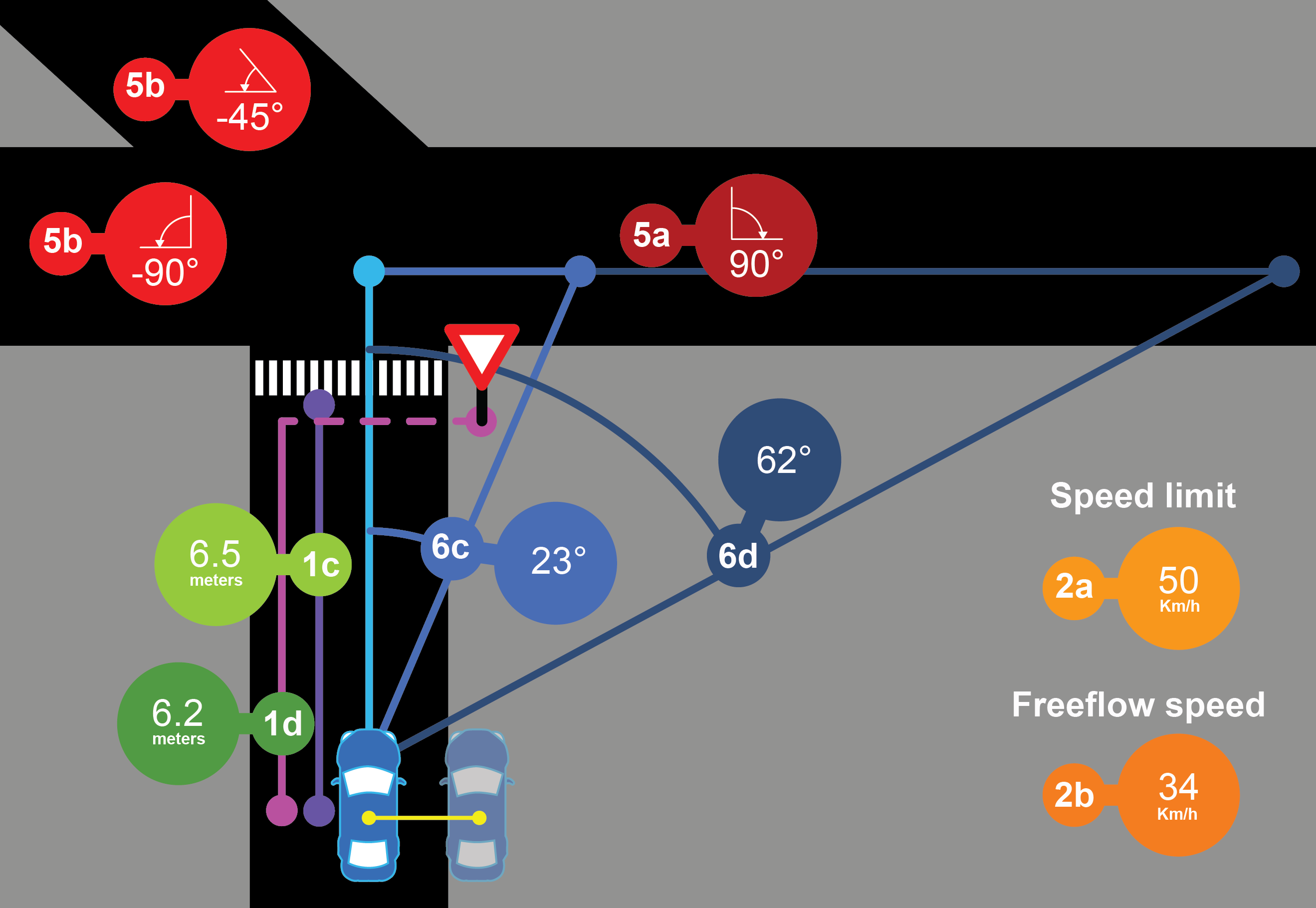}
\caption{An illustration of HERE map features used in this work. Please refer to Table~\ref{table:hereFeatures} for a detailed feature description.} 
\label{fig:here_features}
\end{figure}
\setlength{\textfloatsep}{10pt}
\begin{table*}[tb]
\small
\begin{tabular*}{\textwidth}{|l@{\extracolsep{\fill}}ll|} \hline
Category and Name & Range  & Description \\
\hline
1.a \textit{distanceToIntersection} & $[0m, 250m]$ & \begin{tabular}[c]{@{}l@{}}Road-distance to next intersection encounter. \\ \end{tabular} \\
1.b \textit{distanceToTrafficLight} & $[0m, 250m]$ & Road-distance to next traffic light encounter. \\ 
1.c \textit{distanceToPedestrianCrossing} & $[0m, 250m]$ & Road-distance to next pedestrian crossing encounter. \\ 
1.d \textit{distanceToYieldSign} & $[0m, 250m]$ & Road-distance to next yield sign encounter. \\ \cdashline{1-3}
2.a \textit{speedLimit} & $[0km/h, 120km/h]$ & Legal speed limit for road sector.\\
2.b \textit{freeFlowSpeed} & $[0 km/h, \infty km/h)$ & \begin{tabular}[c]{@{}l@{}}Average driving speed based on underlying road geometry. \end{tabular}\\ \cdashline{1-3}
3.a \textit{curvature} & $[0 m^{-1}, \infty m^{-1})$ & \begin{tabular}[c]{@{}l@{}}Inverse radius of the approximated road geometry by means\end{tabular} \\ \cdashline{1-3}
4.a \textit{turnNumber} & $[0, \infty)$ & \begin{tabular}[c]{@{}l@{}}Index of road at next intersection to travel (counter-clockwise).  \end{tabular} \\ \cdashline{1-3}
5.a \textit{ourRoadHeading} & $[\ang{-180}, \ang{180})$ &  Relative heading of road that car will take at next intersection.\\
5.b \textit{otherRoadsHeading} & $(\ang{-180}, \ang{180})$ &  Relative heading of all other roads at next intersection.\\ \cdashline{1-3}
6.a - 6.e \textit{futureHeadingXm}  & $[\ang{-180}, \ang{180})$ & \begin{tabular}[c]{@{}l@{}} Relative heading of map matched GPS coordinate in   \\ $X\in\{1,5,10,20,50\}$ meters. 
\end{tabular} \\
\hline
\end{tabular*}
\caption{A summary of HERE map data used in this work. }
\label{table:hereFeatures}
\vspace{-5mm}
\end{table*}

We forward the drivelet at $t$ to $\mathbf{G}$ to obtain the driving actions $\hat{B}_t$. To make autonomous driving more human-like is equivalent to letting the distribution of $\hat{B}_t$ approximate that of $B_t$. We thus define our loss for human-like driving as an adversarial loss: 
\begin{equation}
    \label{eq:objective:adverse} 
    \mathcal{L}_t^{hum} = - \text{log}(\mathbf{D}(\hat{B}_t)^{\text{1}}),
\end{equation}
where $\mathbf{D}(\hat{B}_t)^{\text{1}}$ is the probability of classifying $\hat{B}_t$ as human driving. 

Putting everything together, our objective for accurate, comfortable and human-like driving is as follows: 
\begin{equation}
    \label{eq:objective:final} 
    \mathcal{L} =  \sum_{t=1}^{T-O} \sum_{o=1}^O ( \mathcal{L}_t^{\text{acc}} + \zeta_1 \mathcal{L}_t^{\text{com}}) + \zeta_2 \mathcal{L}_t^{hum}. 
\end{equation}
$\zeta_2$ is a trade-off parameter to control the contributions of the costs. 
In keeping with adversarial learning, our training is conducted under the following min-max criterion: 
\begin{equation}
    \label{eq:min-max} 
    \argminA_{\mathbf{G}} \argmaxA_{\mathbf{D}}  \mathcal{L}(I,M).
\end{equation} 




\subsection{HERE Navigation}
\label{sec:here:features}
In this section, we first describe how we augment real-world driving data, specifically the Drive360 dataset~\cite{drive:surroundview:route:planner} with map data from HERE Technologies. Then, we present our feature extraction to translate obtained map data to feature vectors $M$'s in order to be used by our driving model.     

\textbf{Obtaining HERE Map Data}.
Drive360 features $60$ hours of real-world driving data over $3000$ km. We augment Drive360 with HERE Technologies map data. 
Drive360 offers a time stamped GPS trace for each route recorded. We use a path-matcher based on a hidden markov model employing the Viterbi algorithm \cite{forney1973viterbi} to calculate the most likely path traveled by the vehicle during dataset recording, snapping the GPS trace to the underlying road network. This improves our localization accuracy significantly, especially in urban environments where the GPS signal may be weak and noisy. Through the path matcher we obtain a map matched GPS coordinate for each time stamp, which is then used to query the HERE Technologies map database to obtain the various types of map data. 

HERE Technologies has generated an abundant amount of map data. We selected $15$ types of data of $6$ categories, as described in Table \ref{table:hereFeatures}. All features belonging to category 1 will be capped at 250m, for example no $distanceToTrafficLight$ feature is given if the next traffic light on route is further than 250m from the current map matched position. The features of category 5 specify the relative heading of all roads exiting the next approaching intersection, with regard to the map matched entry heading, see Fig.~\ref{fig:here_features}. The features of category 6 specify the relative heading of the planned route a certain distance in advance. This relative heading is only calculated with map matched positions. The relative heading is dependent on the road geometry and the route taken, see Fig.~\ref{fig:here_features}. Using more types of map data constitutes our future work. These augmented map data will be made publicly available.  

\textbf{Deploying Map Data}.
Features belonging to categories 1-4 are denoted as $M[1-4]$. At each time step $t$ we sample $M[1-4]_{[t-k, t]}$ with $k$ set to $2s$ into the past with a step size of $0.1s$. This gives us a feature vector of 160 elements that we subsequently feed into a small LSTM network. We found that an LSTM network yields a better performance than a fully connected network for these feature categories. It is worth noticing that for features belonging to category 1 we supply the inverse distance capped at 1, effectively allowing for a value range of [0, 1]. Features belonging to categories 5 and 6 are denoted as $M[5-6]$. At each time $t$ we sample  $M[5-6]_{t}$, obtaining a feature vector of size 7 that we feed into a small fully connected layer network which works well for these two types of features. The engineered map features will be publicly released. 

\section{Experiments}
\label{sec:experiment} 


\subsection{Implementation Details}
The modules of our method are implemented as follows: 

\noindent 
\textbf{Camera}: Our core model consists of a fine-tuned Resnet34~\cite{resnet} CNN to process sequences of front facing camera images, followed by two regression networks to predict steering wheel angle and vehicle speed. The architecture is similar to the baseline model from~\cite{drive:surroundview:route:planner}. The model in this work requires training with a drivelet of $O$ consecutive instances, each having $K$ frames. It means that $O+K-1$ frames are used for each camera in each optimization step. This leads to memory issues when using multiple surround-view cameras. Thus we choose to proceed with a single front-facing camera for this work.  

 \noindent 
\textbf{TomTom}: Following~\cite{drive:surroundview:route:planner}, a fine-tuned AlexNet \cite{alexnet} is used to process the visual map representation from the TomTom Go App.

\noindent 
\textbf{HERE}: An LSTM nework with 20 hidden states to process $M[1-4]$ and one fully connected network with three layers of size 10 to process $M[5-6]$.

\noindent
\textbf{Comfort}: No extra network is needed. The loss is computed according to Eq.~\ref{eq:confort:driving} and gradients are back propagated to adjust the driving network.  


\noindent 
\textbf{Human-like}: A fully-connected, three-layer discriminator network to model human-like driving. The loss is computed according to Eq.~\ref{eq:objective:adverse} to adjust the driving network.


This in turn allows us to define a total of five DNN models that are composed of combinations of our sub-modules, see Table \ref{table:evaluation}. Each model is trained on the same $50$ hours of training data of the Drive360 dataset. 
We employ a discriminator network $\mathbf{D}$ consisting of three fully connected layers each of size 10 to enforce human-like driving by our models. $\mathbf{D}$ is tasked with classifying maneuvers either as being human or machine created using a binary cross entropy loss.  $\mathbf{D}$ is trained using an Adam optimizer with a learning rate of $10^{-4}$.
We train with a batch size of 16 for one epoch on a Titan X GPU. Training for more epochs does not significantly improve convergence, and thus allows us to limit our maximum model training time to around 26 hours. In terms of parameter values, we set $O$ to 5, $k$ to 3, $\lambda$ to 1, $\zeta_1$ to 0.1 and $\zeta_2$ to 1. A larger value for $O$ and $k$ might lead to better performance but it requires more computational power and GPU memory. Values for the other three parameters are set so that the costs are `calibrated' to the same range. The optimal values can be found by cross-validation if needed.

\begin{table*}[]
\footnotesize 
\setlength\tabcolsep{1.2pt}
\begin{tabular*}{\textwidth}{lccccc @{\extracolsep{\fill}}cccccccccccccc} \toprule
 \multirow{3}{*}{\rot{Model id}} &
 \multirow{3}{*}{\rot{Camera}} & \multirow{3}{*}{\rot{Tomtom}} & \multirow{3}{*}{\rot{HERE}} & \multirow{3}{*}{\rot{Comfort}} & \multirow{3}{*}{\rot{Hu-like}} & \multicolumn{5}{c}{$\mathbb{S}$: the whole } & \multicolumn{3}{|c}{$\mathbb{A}$: traffic light} & \multicolumn{3}{|c}{$\mathbb{B}$: curved } & \multicolumn{3}{|c}{$\mathbb{C}$: approaching} \\
  \multicolumn{6}{r}{}  & \multicolumn{5}{c}{ evaluation set} 
 & \multicolumn{3}{|c}{or pedestrian line } 
 & \multicolumn{3}{|c}{mountain road} 
 & \multicolumn{3}{|c}{intersections}\\
 \multicolumn{6}{r}{} & $A_s$ $\downarrow$ & $A_v$ $\downarrow$ & $C_{lat}$ $\downarrow$ & $C_{lon}$ $\downarrow$ & $H$ $\uparrow$ & $A_s$ $\downarrow$ & $A_v$ $\downarrow$ & $H$ $\uparrow$ & $A_s$ $\downarrow$ & $A_v$ $\downarrow$ & $H$ $\uparrow$ & $A_s$ $\downarrow$ & $A_v$ $\downarrow$ & $H$ $\uparrow$ \\ \midrule


\cite{end:driving:16} & \YES &  &  &  & & 9.81 & 6.50 & 2.92 & 1.46 & 23.30 & 14.39 & 4.97 & 20.77 & 10.99 & 5.12 & 36.36 & 21.15 & 5.41 & 28.31\\
\cite{drive:surroundview:route:planner} & \YES &  \YES &  &  & & 8.67 & 4.92 & 1.60 & 0.76 & 27.24 & 11.94 & 4.20 & 24.02 & 11.10 & 4.85 & 36.23 & 19.69 & 4.49 & 33.34\\ 
Ours & \YES & \YES & \YES & \YES & \YES & \textbf{7.96} & \textbf{4.79} & \textbf{1.46} & \textbf{0.34} & \textbf{29.31} & \textbf{12.68} & \textbf{3.62} & \textbf{31.56} & \textbf{8.39} & \textbf{3.46} & \textbf{46.64} & \textbf{17.12} & \textbf{4.38} & \textbf{35.68}\\ 
\midrule

1 & \YES &  &  &  & & 10.20 & 5.84 & 2.29 & 0.91 & 20.20 & 13.24 & 4.40 & 20.85 & 9.04 & 4.81 & 35.86 & 21.37 & 4.99 & 28.34\\



2 & \YES & \YES &  & &  & 8.67 & 4.92 & 1.60 & 0.76 & 27.24 & 11.94 & 4.20 & 24.02 & 11.10 & 4.85 & 36.23 & 19.69 & 4.49 & 33.34\\ 
3 & \YES & \YES & \YES &  & & 8.41 & 4.81 & 1.69 & 0.78 & 27.72 & \textbf{10.22} & 3.75 & 28.72 & 9.33 & 4.24 & 38.62 & \textbf{16.83} & 4.40 & \textbf{37.21}\\ 
4 & \YES & \YES  & \YES & \YES & & 8.81 & 5.24 & 1.66 & 0.41 & 27.75 & 11.72 & 4.51 & 27.87 & 12.65 & 4.99 & 28.49 & 18.19 & 4.85 & 35.42\\
5 & \YES & \YES & \YES & \YES & \YES & \textbf{7.96} & \textbf{4.79} & \textbf{1.46} & \textbf{0.34} & \textbf{29.31} & 12.68 & \textbf{3.62} & \textbf{31.56} & \textbf{8.39} & \textbf{3.46} & \textbf{46.64} & 17.12 & \textbf{4.38} & 35.68\\

\bottomrule
\end{tabular*}
\caption{The performance of all variants of our method evaluated on the four evaluation sets defined. Driving accuracy is denoted by $A_v$ \& $A_s$ for speed (km/h) and steering angle (degree), comfort measure by $C_{lat}$ \& $C_{lon}$ for latitude and longitudinal, and the human-likeliness score by $H$ (\%). $\uparrow$ means that higher is better and $\downarrow$ the opposite.}
\label{table:evaluation}
\vspace{-4mm}
\end{table*}

\subsection{Evaluation}
A driving model should drive as accurately as possible in a wide range of scenarios. As our models are trained via imitation learning, we define accuracy as how close the model predictions are to the human ground truth maneuver under the L1 distance metric. We define $A_s$ as the absolute mean error in the steering angle prediction and $A_v$ as the absolute mean error in the vehicle speed prediction. Specifically, we predict the steering wheel angle $\mathcal{S}_{t+0.5s}$ and vehicle speed $\mathcal{V}_{t+0.5s}$ 0.5s into the future \footnote{Predicting further into the future is possible and our experiments have shown a growing degradation in accuracy the further one predicts.}. We use a SmoothL1 loss to jointly train $\mathcal{S}_{t+0.5s}$ and $\mathcal{V}_{t+0.5s}$ using the Adam Optimizer with an initial learning rate of $10^{-4}$.

\textbf{Evaluation Sets}. 
Our whole test set, denoted by $\mathbb{S}$, consisting of around 10 hours of driving, covers a wide range of situations including city and countryside driving. While one overall number on the whole set is easier to follow, evaluations on specific scenarios can highlight the strengths and weaknesses of driving models at a finer granularity.
By enriching the Drive360 dataset with HERE map data, we can filter our test set $\mathbb{S}$ for specific scenarios. We have chosen three interesting scenarios in this evaluation\footnote{more will be include in the supplementary material.}:
\vspace{-1mm}
\begin{itemize}
\itemsep -0.5\parsep
\item $\mathbb{A} \subset \mathbb{S}$ where the distance to the next traffic light is less than 40m or the distance to the next pedestrian crossing is less than 40m and the speed limit is less than or equal to 50km/h. Translates to \textit{approaching a traffic light or pedestrian crossing in the city}.
\item $\mathbb{B} \subset \mathbb{S}$ where the curvature is greater than 0.01 and the speed limit is 80 km/h and the distance to the next intersection greater than 100m. Translates to \textit{winding road where road radius is less than 100m and no intersections in the vicinity}.
\item $\mathbb{C} \subset \mathbb{S}$ where the distance to the next intersection is less than 20m, named \textit{approaching an intersection}. 
\end{itemize}

\textbf{Comparison to state-of-the-art}. 
We compare our method to two start-of-the-art end-to-end driving methods~\cite{end:driving:16} and \cite{drive:surroundview:route:planner}. They are trained under the same settings as our method is trained.
The results are shown at the top of Table~\ref{table:evaluation}. The results show that our method performs better than the two competing methods. It is more accurate, more comfortable and behaves more like a human driver than previous methods. This is due to our numerical map features, the sequence-based learning, and two new learning objectives, namely \emph{ride comfort} and \emph{human-like style}. We also compare our method to the classical proportional–integral–derivative (PID) controller for improving ride comfort and show that our method performs better (see Table~\ref{table:PID_evaluation}). All detailed discussions are given in the following sections, where we also compare against ablations of our full model (id: 5).

\subsubsection{Driving Accuracy}

By comparing the performance of $Model_{1-3}$ in Table~\ref{table:evaluation}, one can find that driving accuracy improves significantly when using maps. $Model_1$ and $Model_2$ are in fact the same models as used in~\cite{drive:surroundview:route:planner}. The best results are achieved by using HERE numerical features and TomTom visual maps together. This implies that the two ways of using map data are to some extent complementary. 
We reason that the TomTom module offers a complete world view, in other words, an aggregation of all available TomTom map data, rendered into a video. It is designed to facilitate human driving, but it seems that neural networks benefit from having this intuitive representation as well. Yet, while the rendered video is quite effective for representing route information, there is little room left for further improvement towards accurate navigation -- it is very challenging to reverse engineer all exact map information from the rendered videos. 

The designed features out of our numerical map data are in stark contrast to the rendered video representation. They are accurate and unequivocal. By using numerical map features, our method $Model_3$ outperforms \cite{drive:surroundview:route:planner} which only uses rendered visual maps. It also outperforms \cite{end:driving:16} which uses no map information.  Our engineered HERE map features show marked improvement for challenging driving scenarios.  For instance, the error (mse) of steering angle prediction is reduced from $19.69$ to $16.83$ for scenario \emph{approaching intersections} by using our engineered map features, and the error for speed prediction is reduced from $4.20$ to $3.75$ for \emph{traffic light or pedestrian crossing}. In this work, only fifteen numerical features are hand selected from the vast amount of available HERE map features. Logically, this leaves plenty of room for improving the use of numerical map features. This is especially true as the quality of navigation maps keeps improving. For rendered videos, one has almost exhausted the ability for further improvement. Part of our future work will be on incorporating a much larger set of numerical map features. 

\subsubsection{Ride Comfort}

By imposing an additional comfort loss, we are able to significantly improve ride comfort by reducing lateral and longitudinal oscillations, at a modest loss of driving accuracy. This can be confirmed by comparing the performance of $Model_3$ and $Model_4$ in Table~\ref{table:evaluation}.  
For models that use no map data, we observe a similar trend -- significant gains in comfort at a modest cost in driving accuracy. One direction to address this issue is to design and learn an adaptive loss for \emph{ride comfort} such that it only takes effect when the driving model is accurate. That is to say, \emph{ride comfort} measures are only applied when driving scenarios are easy. Otherwise, safety considerations must take prevalence. Using road attributes from our map features can help identify such scenarios. We have also investigated in how far a classical proportional–integral–derivative (PID) controller can achieve similar comfort levels to our learned approach. To this end, we process the network predictions of \cite{end:driving:16} and \cite{drive:surroundview:route:planner} with tuned (exhaustive grid search for parameters) PID controllers such that their comfort score is the same as our learned approach. We report these results in Table~\ref{table:PID_evaluation}. This shows that reaching similar levels of comfort to our learned approach with a PID controller comes at the price of degraded driving performance and thus our way of including comfort in the driving model is preferable.

\begin{table}[]
\footnotesize
\setlength\tabcolsep{1.2pt}
\begin{tabular*}{\columnwidth}{l @{\extracolsep{\fill}}ccccc} \toprule
 Method & $A_s$ $\downarrow$ & $A_v$ $\downarrow$ & $C_{lat}$ $\downarrow$ & $C_{lon}$ $\downarrow$ & $H$ $\uparrow$ \\ \midrule
\cite{end:driving:16}+PID & 10.72 & 6.76 & \textbf{1.46} & \textbf{0.34} & 22.73  \\
\cite{drive:surroundview:route:planner}+PID & 8.81 & 4.85 & \textbf{1.46} & \textbf{0.34} & 27.56 \\
Ours & \textbf{7.96} & \textbf{4.79} & \textbf{1.46} & \textbf{0.34} & \textbf{29.31} \\
\bottomrule
\end{tabular*}
\caption{The performance of \cite{end:driving:16} and \cite{drive:surroundview:route:planner} with an additional PID controller evaluated on the full set $\mathbb{S}$. To be read like Table \ref{table:evaluation}}
\label{table:PID_evaluation}
\end{table}

\setlength{\textfloatsep}{9pt}
\begin{figure}
    \centering
    \begin{tikzpicture}[scale=0.95] 
    \tikzstyle{every node}=[font=\small]
    \begin{axis}[
            ybar,
            ymode=linear,
            width=\textwidth/2,
            height=5cm,
            xmin=0,
            xmax=20,
            ymin=0,
            ymax=100,
            ylabel={Relative Error Rate (\%)},
            xtick={2.5, 6.5, 9.5, 13, 17.5},
            minor xtick={5,8,11,15},
            xticklabels = {
                Speed Limit (km/h),
                Traffic Light,
                Cross Walk,
                Road Type,
                Intersection
            },
            x tick label style={anchor=center,
            font=\tiny},
            major x tick style = {opacity=0},
            minor x tick num = 1,
            xtick pos=left,
            ymajorgrids=true,
            every node near coord/.append style={
                    anchor=west,
                    rotate=90,
                    font=\tiny,
            }
            ]

    \addplot[bar shift=0pt,draw=30,          fill opacity=0.9,fill=30!80!white           , nodes near coords=30                 ] plot coordinates{ ( 1,     \nangthirty  ) };
    \addplot[bar shift=0pt,draw=50,      fill opacity=0.9,fill=50!80!white       , nodes near coords=50              ] plot coordinates{ ( 2,     \nangfifty   ) };
    \addplot[bar shift=0pt,draw=80,      fill opacity=0.9,fill=80!80!white       , nodes near coords=80              ] plot coordinates{ ( 3,     \nangeighty   ) };
    \addplot[bar shift=0pt,draw=120,      fill opacity=0.9,fill=120!80!white       , nodes near coords=120              ] plot coordinates{ ( 4,     \nangonetwenty   ) };

    \addplot[bar shift=0pt,draw=Light Near,      fill opacity=0.9,fill=Light Near!80!white       , nodes near coords=Yes               ] plot coordinates{ ( 6,     \nangtlnear  ) };
    \addplot[bar shift=0pt,draw=Light Far,         fill opacity=0.9,fill=Light Far!80!white          , nodes near coords=No              ] plot coordinates{ ( 7,     \nangtlfar    ) };

    \addplot[bar shift=0pt,draw=Crossing Near,    fill opacity=0.9,fill=Crossing Near!80!white     , nodes near coords=Yes               ] plot coordinates{ ( 9,    \nangcwnear  ) };
    \addplot[bar shift=0pt,draw=Crossing Far,       fill opacity=0.9,fill=Crossing Far!80!white        , nodes near coords=No              ] plot coordinates{ ( 10,    \nangcwfar   ) };

    \addplot[bar shift=0pt,draw=Left,           fill opacity=0.9,fill=Left!80!white            , nodes near coords=Left Bend           ] plot coordinates{ ( 12,    \nangleft   ) };
    \addplot[bar shift=0pt,draw=Straight,         fill opacity=0.9,fill=Straight!80!white          , nodes near coords=Straight         ] plot coordinates{ ( 13,    \nangstraight    ) };
    \addplot[bar shift=0pt,draw=Right,         fill opacity=0.9,fill=Right!80!white          , nodes near coords=Right Bend        ] plot coordinates{ ( 14,    \nangright    ) };
    \addplot[bar shift=0pt, draw=Right, white!20!white, opacity=0.0, fill opacity=0.0, fill=white!80!white] plot coordinates{ ( 15,  85   ) };
    
    \addplot[bar shift=0pt,draw=Enter,           fill opacity=0.9,fill=Enter!80!white            , nodes near coords=Approach                  ] plot coordinates{ ( 16,    \nangenter  ) };
    \addplot[bar shift=0pt,draw=Inside,          fill opacity=0.9,fill=Inside!80!white           , nodes near coords=Inside                 ] plot coordinates{ ( 17,    \nanginside  ) };
    \addplot[bar shift=0pt,draw=Exit,  fill opacity=0.9,fill=Exit!80!white   , nodes near coords=Depart         ] plot coordinates{ ( 18,    \nangexit    ) };
    \addplot[bar shift=0pt,draw=None, fill opacity=0.9,fill=None!80!white  , nodes near coords=None       ] plot coordinates{ ( 19,   \nangnone   ) };
    \node[above,font=\small\bfseries] at (current bounding box.north) {Error for Steering Angle};
    \end{axis}
    \end{tikzpicture}
    \qquad
    \begin{tikzpicture}[scale=0.95] 
    \tikzstyle{every node}=[font=\small]
    \begin{axis}[
            ybar,
            ymode=linear,
            width=\textwidth/2,
            height=5cm,
            xmin=0,
            xmax=20,
            ymin=0,
            ymax=100,
            ylabel={Relative Error Rate (\%)},
            xtick={2.5, 6.5, 9.5, 13, 17.5},
            minor xtick={5,8,11,15},
            xticklabels = {
                Speed Limit (km/h),
                Traffic Light,
                Cross Walk,
                Road Type,
                Intersection
            },
            x tick label style={anchor=center,
            font=\tiny},
            major x tick style = {opacity=0},
            minor x tick num = 1,
            xtick pos=left,
            ymajorgrids=true,
            every node near coord/.append style={
                    anchor=west,
                    rotate=90,
                    font=\tiny,
            }
            ]

    \addplot[bar shift=0pt,draw=30,          fill opacity=0.9,fill=30!80!white           , nodes near coords=30                 ] plot coordinates{ ( 1,     \nvelthirty ) };
    \addplot[bar shift=0pt,draw=50,      fill opacity=0.9,fill=50!80!white       , nodes near coords=50              ] plot coordinates{ ( 2,     \nvelfifty   ) };
    \addplot[bar shift=0pt,draw=80,      fill opacity=0.9,fill=80!80!white       , nodes near coords=80              ] plot coordinates{ ( 3,     \nveleighty   ) };
    \addplot[bar shift=0pt,draw=120,      fill opacity=0.9,fill=120!80!white       , nodes near coords=120              ] plot coordinates{ ( 4,     \nvelonetwenty   ) };

    \addplot[bar shift=0pt,draw=Light Near,      fill opacity=0.9,fill=Light Near!80!white       , nodes near coords=Yes               ] plot coordinates{ ( 6,     \nveltlnear  ) };
    \addplot[bar shift=0pt,draw=Light Far,         fill opacity=0.9,fill=Light Far!80!white          , nodes near coords=No               ] plot coordinates{ ( 7,     \nveltlfar    ) };

    \addplot[bar shift=0pt,draw=Crossing Near,    fill opacity=0.9,fill=Crossing Near!80!white     , nodes near coords=Yes               ] plot coordinates{ ( 9,   \nvelcwnear  ) };
    \addplot[bar shift=0pt,draw=Crossing Far,       fill opacity=0.9,fill=Crossing Far!80!white        , nodes near coords=No              ] plot coordinates{ ( 10,    \nvelcwfar   ) };

    \addplot[bar shift=0pt,draw=Left,           fill opacity=0.9,fill=Left!80!white            , nodes near coords=Left Bend           ] plot coordinates{ ( 12,    \nvelleft ) };
    \addplot[bar shift=0pt,draw=Straight,         fill opacity=0.9,fill=Straight!80!white          , nodes near coords=Straight         ] plot coordinates{ ( 13,    \nvelstraight   ) };
    \addplot[bar shift=0pt,draw=Right,         fill opacity=0.9,fill=Right!80!white          , nodes near coords=Right Bend        ] plot coordinates{ ( 14,    \nvelright    ) };
    \addplot[bar shift=0pt, draw=Right, white!20!white, opacity=0.0, fill opacity=0.0, fill=white!80!white] plot coordinates{ ( 15,  85   ) };
    
    \addplot[bar shift=0pt,draw=Enter,           fill opacity=0.9,fill=Enter!80!white            , nodes near coords=Approach                  ] plot coordinates{ ( 16,    \nvelenter  ) };
    \addplot[bar shift=0pt,draw=Inside,          fill opacity=0.9,fill=Inside!80!white           , nodes near coords=Inside                 ] plot coordinates{ ( 17,    \nvelinside   ) };
    \addplot[bar shift=0pt,draw=Exit,  fill opacity=0.9,fill=Exit!80!white   , nodes near coords=Depart         ] plot coordinates{ ( 18,    \nvelexit  ) };
    \addplot[bar shift=0pt,draw=None, fill opacity=0.9,fill=None!80!white  , nodes near coords=None       ] plot coordinates{ ( 19,   \nvelnone   ) };
    \node[above,font=\small\bfseries] at (current bounding box.north) {Error for Speed Control};
    \end{axis}
    \end{tikzpicture}
    \vspace{-2mm}

    \caption{Relative Error Rate (\%) for five road attributes. }
    \label{fig:relative_error_rate}
\end{figure}
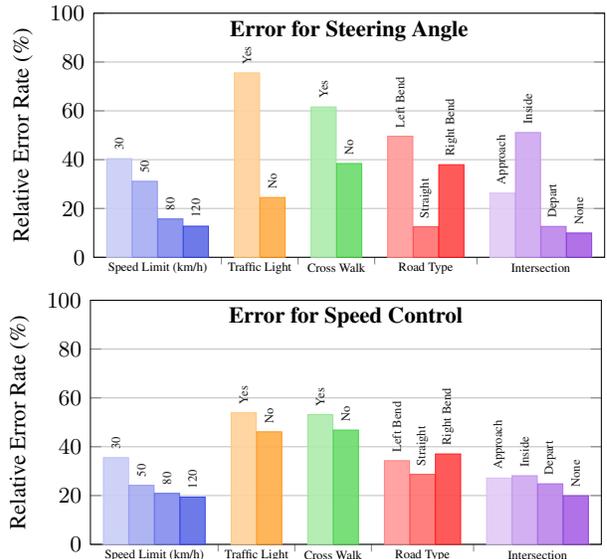
\vspace{-2mm}

\setlength{\textfloatsep}{10pt}
\setlength{\tabcolsep}{1pt}
\begin{figure*}[!tb]
$\begin{tabular}{cccccc}
\text{ (1)} &
\includegraphics[width=0.30\linewidth]{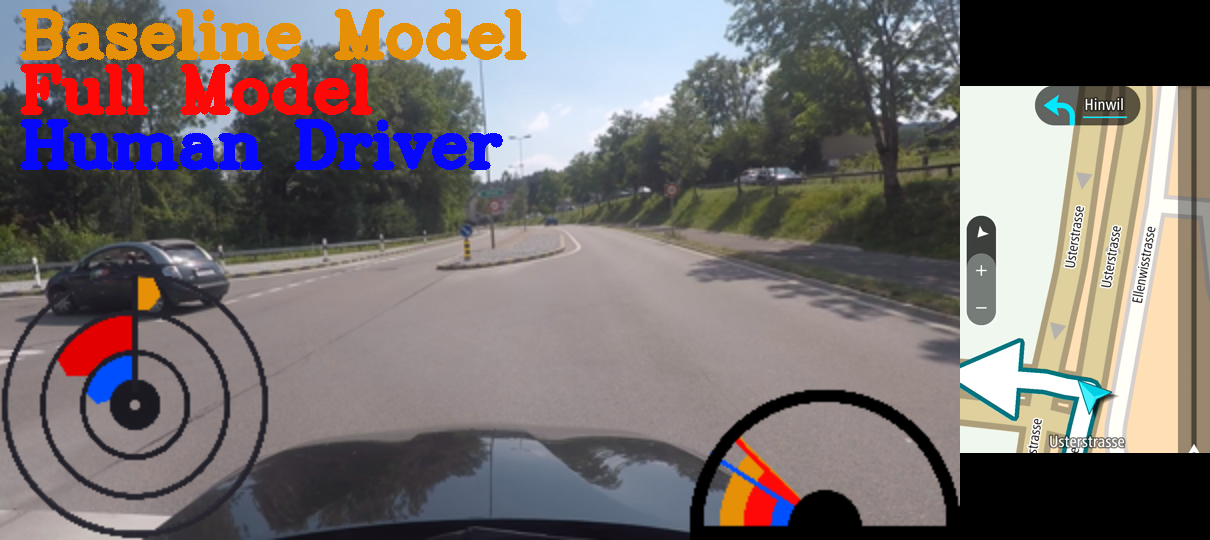}
& 
\includegraphics[width=0.30\linewidth]{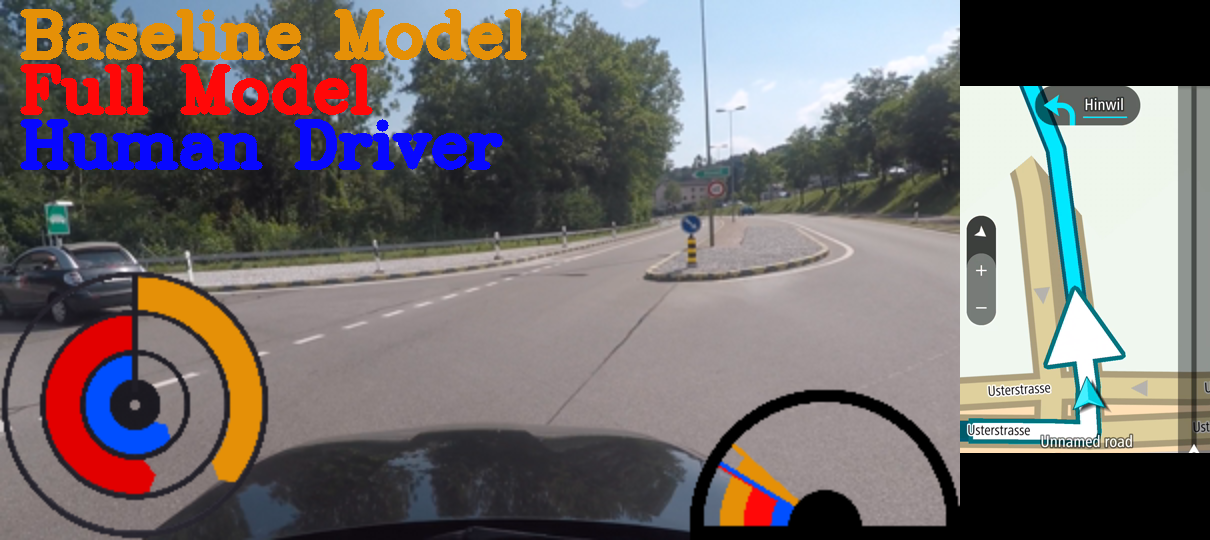}
& 
\includegraphics[width=0.30\linewidth]{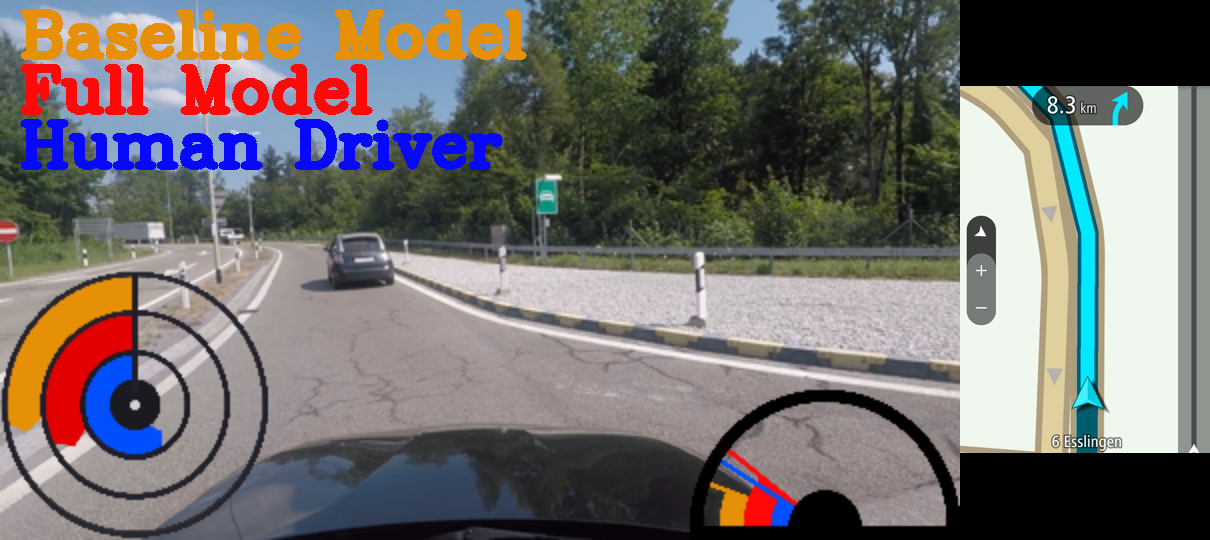} \\

\text{ (2)} &
\includegraphics[width=0.30\linewidth]{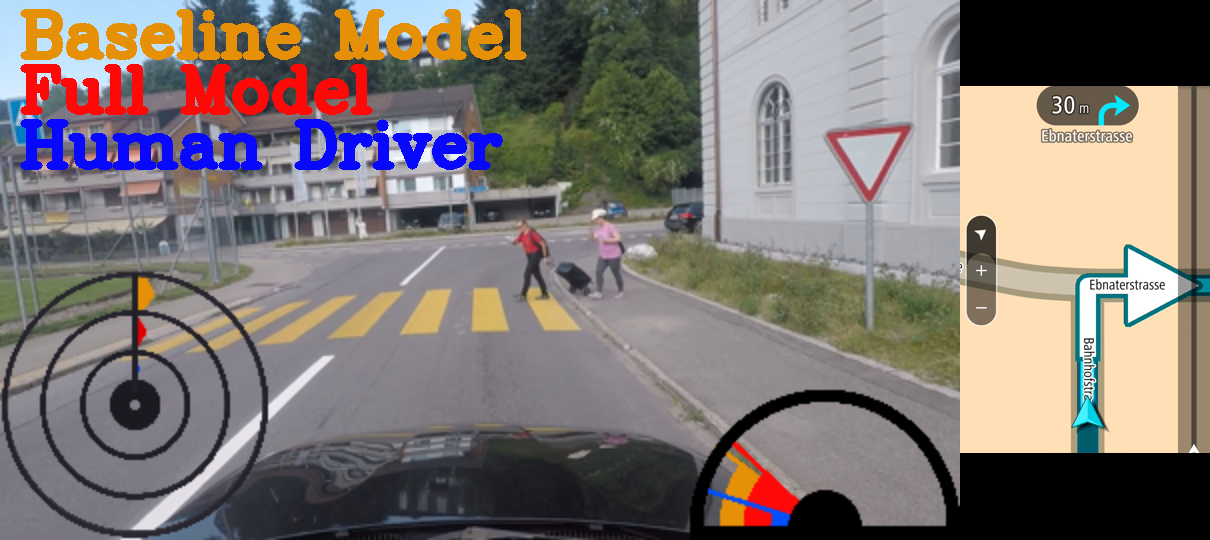}
& 
\includegraphics[width=0.30\linewidth]{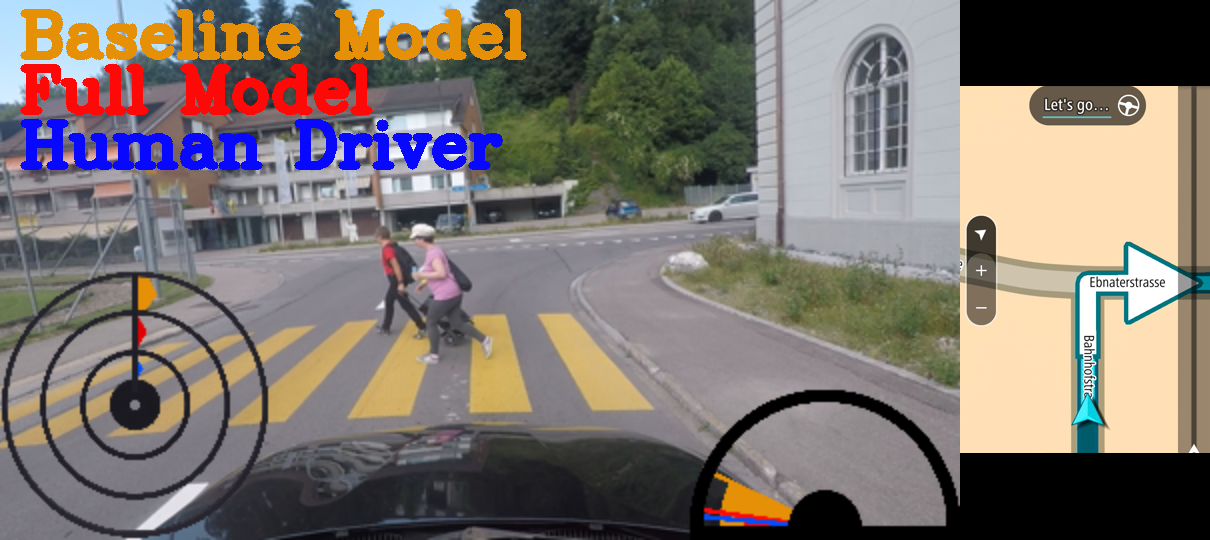} & 
\includegraphics[width=0.30\linewidth]{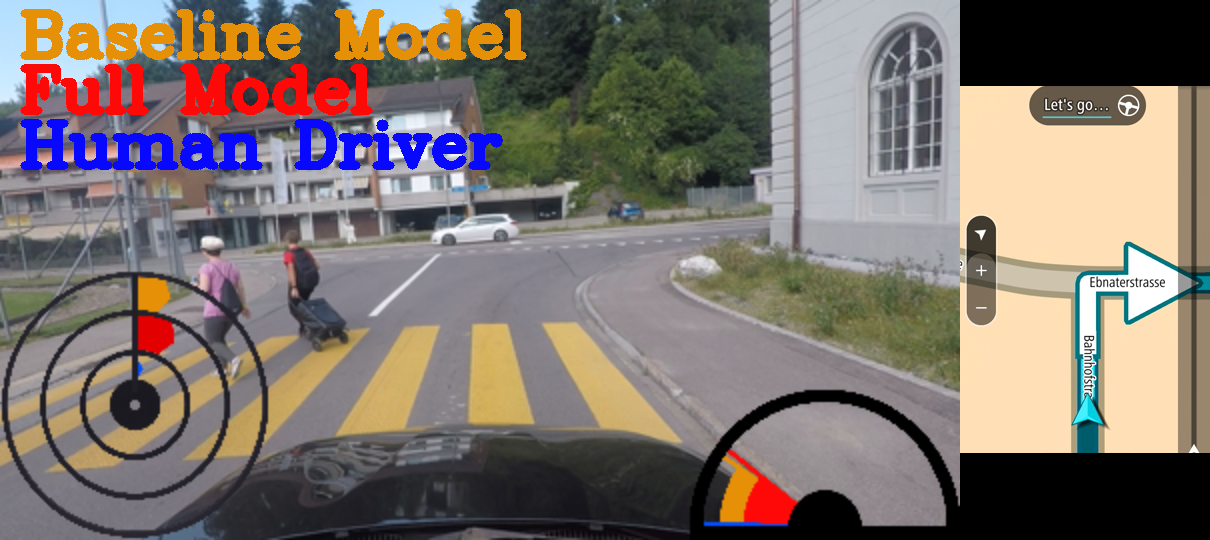}  \\
& \text{(a) $t$ } &  \text{ (b) $t$ +1s } & \text{ (c) $t$ +2s } \\ 
\end{tabular}$
 \caption{Qualitative results for future driving action prediction, to compare our full driving model (id: 5) to the baseline of front camera-only-model (id: 1). Decisions at three time steps over around $2$ seconds are shown for two driving scenarios. Better seen on screen.} 
\label{fig:examples:pics} 
\end{figure*}
\vspace{-2mm}

\subsubsection{Human-likeness}
A driving style to be human-like is hard to quantify. It is also hard to evaluate. In order to evaluate it quanlitatively, 
we propose a new evaluation criterion -- the \textit{human-likeness score}. This score is calculated by clustering human driving maneuvers ($s$ and $v$ concatenated) from the evaluation set $\mathbb{S}$, over a $0.5$s window with a step size of $0.1$s, into 75 unique clusters using the Kmeans algorithm. Predicted model maneuvers are then considered human-like if, for the same window, they are associated with the same cluster as the human maneuver. We chose our window and step size to be consistent with our model training. The \textit{human-likeness score} is then defined as the percentage of driving maneuvers given by an algorithm being associated to the same cluster as the human driving maneuvers for the same time window. 
To this end, and similar to the comfort training procedure, we generate model driving maneuvers via a sequence of five ($O=5$) consecutive steering wheel angle and vehicle speed predictions.
We observe, again in Table~\ref{table:evaluation}, that our adversarial learning designed for modeling a human-like driving style significantly improves overall performance and in particular boosts driving accuracy and the human-likeness score, see $Model_{5}$ vs $Model_{4}$. 
Interestingly, when a model drives more accurately, due to the presence of a navigation component, its human-likeness score improves as well. This is evidenced by the performance of $Model_1$ and $Model_{3}$. This is because the model has a clearer understanding of the driving environment and consequently yields quite comfortable and human-like driving already. 
Overall, the model trained using our human likeness loss, along with the accuracy loss and ride comfort loss, drives more accurate, more comfortable and more human-like than all previous methods.

\subsubsection{Error Diagnosis} 
It is notoriously difficult to evaluate and understand driving model performance solely based on a global quantitative evaluation. For example, a model with the lowest overall test steering error might perform terribly in certain specific scenarios. 
The ability to automatically generate meaningful evaluation subsets using numerical map features is tremendously helpful and allows for scenario targeted evaluations as demonstrated in Table~\ref{table:evaluation}.  In this section, we propose a new evaluation scheme to visualize the correlation of the error rate (frequency of making errors) of our final model to road attributes. 
More specifically, five major road attributes (speed limit, traffic light, cross walk, road type, and intersection) are chosen; for each of them, we define exclusive subsets that collectively complete the test set. We then calculate the error-rate for each subset, where, as in~\cite{driving:failure:prediction}, error is defined when steering prediction is off by more than $10$ $\deg$ or speed prediction by more than $5$ $km/h$. For each road attribute, we normalize the error-rates over its subsets. 
The results are shown in Fig.~\ref{fig:relative_error_rate}. As can be seen, the model generally makes more mistakes in more challenging situations. For instance, it makes more mistakes on winding roads than on straight roads, and makes more mistakes at intersections than along less taxing stretches of road. We can also see that the model works better on highways than in cities. The results of this error diagnosis provide new insights for the further improvement of existing methods and is therefore advisable to include.  

\subsubsection{Qualitative Evaluation}
In Fig.~\ref{fig:examples:pics} we present two unique driving sequences with respective model predictions. The steering wheel angle gauge, on the left, is a direct map of the steering wheel angle to degrees, whereas the speed gauge, on the right, is from 0km/h to 130km/h. Gauges should be used for relative model comparison, with our baseline $Model_1$ prediction in orange, our most accurate $Model_{5}$ prediction in red and the human maneuver in blue. We consider a model to perform well when the magnitude of a gauge is identical (or similar) to the human gauge.

Fig.~\ref{fig:examples:pics} (1) shows the benefit of using maps. $Model_{1}$ falsely predicts continuing straight on the road and tries to compensate for leaving the lane, see (3,b), while $Model_{5}$ accurately predicts the left turn. By using this example, we would like to emphasize again that due attention should be paid to map data when learning driving models. Fig.~\ref{fig:examples:pics} (2) shows the benefit of using HERE features. While one may claim that $Model_{5}$, being aware of distance to pedestrian crossings, accurately slowed down for the pedestrians in (2, b) and would, in a real setting not have had an accident, $Model_{1}$ clearly did no such thing. Thus a great benefit of HERE features is that we can automatically filter for these scenarios and evaluate models at a finer granularity. This greatly improves model understanding. Please see the supplementary material for a video of animated sequences.

\section{Conclusion}
This paper has extended the objective of autonomous driving models from accurate driving to accurate, comfortable and human-like driving. The importance of the three objectives has been thoroughly discussed, mathematically formulated, and then translated into one neural network which is end-to-end trainable. This work made three contributions. First, numerical maps from the leading mapping company HERE Technologies are employed to augment the $3000$km real-world driving data of the Drive360 dataset. A set of driving-relevant features have been extracted to effectively use the map information for autonomous driving. Second, the learning of end-to-end driving models is improved from pointwise prediction to sequence-based prediction and a passengers' comfort measure is included to reduce motion sickness. Finally, adversary learning was introduced such that the learned driving model behaves more like human drivers.  Extensive experiments have shown that our driving model is more accurate, more comfortable and more human-like than previous methods. 

\textbf{Acknowledgement} 
This work is funded by Toyota Motor Europe via the research
project TRACE-Zurich. We are grateful for the support by HERE Technologies for granting us the access of their map data, without their support this project would not have been possible. We would like to thank in particular Dr. Harvinder Singh for his insightful discussions of using HERE map data. 

{\small
\bibliographystyle{ieee}
\bibliography{egbib}
}

\end{document}